\documentclass{article}

\PassOptionsToPackage{numbers,compress}{natbib}
\usepackage[final,dblblindworkshop]{neurips_2026}
\workshoptitle{Self-Evolving Diversity-Driven Search for Robust AI Systems}

\usepackage[utf8]{inputenc}
\usepackage[T1]{fontenc}
\usepackage{hyperref}
\usepackage{url}
\usepackage{booktabs}
\usepackage{amsfonts}
\usepackage{amsmath}
\usepackage{nicefrac}
\usepackage{microtype}
\usepackage{xcolor}
\usepackage{graphicx}
\usepackage{array}

\newcommand{\taubench}{$\tau^2$-bench}

\title{Generated Evaluation Suites as Diversity-Driven Search for Real Agent Regressions}

\author{
  Pulkit Yadav \\
  Independent Researcher \\
  \texttt{pulkityadav1@gmail.com}
}

\begin{document}

\maketitle

\begin{abstract}
Generating an evaluation suite from an agent's specification is diversity-driven search: a generator fills a descriptor space (requirements $\times$ personas $\times$ scenario kinds), and the suite it emits is an archive, then scored like a benchmark by its mean pass rate, which throws away the search's product. We inject 7 controlled defects into a \taubench{} agent across two domains. Coverage yields a ceiling: if a defect's traced footprint holds $b$ scenarios that pass at baseline, it can itself contribute at most $b/N$ to a suite of $N$. For 4 of 5 traceable defects that ceiling lies at or below the detection threshold $\theta = 0.10$, so no within-footprint effect can move the suite past it; in both domains the tool-removal defect destroys every covered scenario it can (exactly its ceiling) while the suite-level pass rate moves the other way. Defects split into two regimes: trace-aligned ones, where per-cell scoring recovers the tool-removal defect in full and reveals a second as a significant improvement, and trace-orthogonal ones, whose footprint is behavioural: forced escalation is detected in retail at a traced coverage of exactly zero. Diversity along the generator's own axes helps only where it aligns: at fixed suite size a happy-path-only suite detects nothing, while a kind-diverse one detects the orchestration defect in 5 of 20 draws. Evaluation archives should be scored per cell and should publish coverage as a first-class property.
\end{abstract}

\section{Introduction}

An agent's specification (policy, tool schemas, routing rules, prompts) changes weekly while its evaluation suite is written once. A widely adopted response is to \emph{generate} the suite from it: extract requirements, generate personas, cross them into scenarios, grade against predicates~\citep{katharki2026goal}. Read as search this is a familiar object: the generator fills a descriptor space, the suite is viewed as a descriptor-indexed archive~\citep{mouret2015mapelites,pugh2016qd}, and the argument for it is the argument for divergent search: coverage beats optimisation of a single objective~\citep{lehman2011novelty}.

The archive is then scored like a benchmark: run every scenario, average, compare. The mean over a diverse archive is dominated by the cells where nothing interesting happened. We ask what that mismatch costs when the question is practical (\emph{did this suite notice that the agent regressed?}), answering it with mutation testing~\citep{demillo1978hints,jia2011mutation} over controlled defects injected into a \taubench{} agent~\citep{yao2024taubench,barres2025tau2}. The answer is not that the search is bad, but that coverage and score are misaligned in a way computable from baseline results alone, before the defect is introduced: for most defects the ceiling on the defect-attributable signal already sits below the detection threshold.

\textbf{Contributions.} We formulate generated suites as diversity-driven search and give a coverage measure over the generator's own descriptors, with a ceiling bounding what suite-level scoring can detect (\S\ref{sec:coverage}); show over 7 defects and two domains that the ceiling binds for 4 of 5 traceable defects, and that a defect can saturate its ceiling while the suite-level pass rate moves the opposite way (\S\ref{sec:ceiling}); separate trace-aligned from trace-orthogonal defects, each demanding a different unit of scoring (\S\ref{sec:regimes}); and report two diversity ablations that locate where diversity does and does not buy detection (\S\ref{sec:diversity}). We will release the pipeline, mutation set, suites and transcripts.

\section{Suites as archives}

\textbf{Diversity-driven search.} The archive formulation above comes from novelty search, MAP-Elites and quality-diversity~\citep{lehman2011novelty,mouret2015mapelites,pugh2016qd}. The idea reaches language models mainly through adversarial generation: Rainbow Teaming casts red-teaming as quality-diversity over a descriptor grid~\citep{samvelyan2024rainbow,perez2022redteaming}. Those methods search for \emph{inputs that break a fixed model}; we study the dual, a fixed suite asked to notice when the \emph{system} changes underneath it.

\textbf{Agents, benchmarks, mutation analysis.} $\tau$-bench, \taubench{}~\citep{yao2024taubench,barres2025tau2} and AgentBench~\citep{liu2024agentbench} score agents on fixed task sets that are easily overfit~\citep{kapoor2024agents}; MAST~\citep{cemri2025mast} and issue-tracker mining~\citep{zhu2026where,asgari2025challenges,zhang2026bugs} catalogue the failures that dominate in deployment, and our mutation classes come from them. Specification-grounded generation~\citep{katharki2026goal} composes LLM test generation~\citep{kang2023fewshot}, simulated users~\citep{schatzmann2007agenda} and LLM judges~\citep{zheng2023judging}. Mutation analysis scores a suite by the seeded faults it kills~\citep{demillo1978hints,jia2011mutation}; here the mutated object is an \emph{agent configuration} and the tests are stochastic conversations, so a kill must be defined statistically, and that definition, we show, can constrain detection before mutant evaluation.

\section{Method}

\subsection{The archive}

The generator consumes a specification $S$ (for \taubench{} v1.0.1 (fc0055d), the domain policy document) plus tool schemas and emits executable scenarios with grading predicates over three descriptors: the \emph{requirement} exercised, the \emph{persona} driving the conversation, and the scenario \emph{kind} (happy path, edge case, adversarial-benign). In both domains it extracts 30 requirements, writes 8 personas, and emits 3 scenarios per requirement, one per kind: a fully covered $30 \times 3$ grid of 90 scenarios. All 90 run on airline; retail evaluates a seeded 50 of them, and its S3 deltas are over those 50. Simulation runs in the benchmark's unmodified harness. Models: \texttt{gpt-5.6-luna} as agent, user simulator and judge, \texttt{gpt-5.6-terra} as generator; temperature is left at the provider default because the agent model rejects temperature 0, and reasoning is disabled (zero reasoning tokens in the usage records). All runs took place on 29 August 2026 (Appendix~\ref{app:repro}).

\subsection{The defect space}

Each mutation is a minimal, realistic edit to the agent configuration that leaves the harness untouched, modelling a failure class documented in the mining studies above. We instantiate 7 in the airline domain, one per operator: \textbf{R1} deletes a policy clause (the 24-hour cancellation rule), \textbf{R2} inverts a policy condition, \textbf{T1} removes a tool (\texttt{cancel\_reservation}), \textbf{T2} renames a tool-schema parameter, \textbf{O1} inverts the escalation rule so the agent always transfers to a human, and \textbf{C1} and \textbf{C2} truncate the policy to its first 40\% and history to the last 4 turns (Appendix~\ref{app:mutations}). This is a pilot-scale set: each result is a statement about one instance, not its operator class.

\subsection{Scoring, coverage, and the ceiling}
\label{sec:coverage}

Let $p(\Sigma, A)$ be the pass rate of agent $A$ on suite $\Sigma$ and $A_m$ be $A$ with mutation $m$. A generated scenario passes on its programmatic end-state check where one exists (61 of 90 on airline) and on the judge's verdict otherwise; hand-authored tasks pass on the benchmark's own checker. \textbf{Score, specified before running experiments.} $\Sigma$ \emph{detects} $m$ if $\Delta_m = p(\Sigma, A) - p(\Sigma, A_m)$ exceeds $\theta = 0.10$ and a 2,000-sample paired bootstrap over scenarios yields a 95\% interval excluding zero. \textbf{Why $\theta = 0.10$.} We fixed $\theta$ in the experiment plan, together with the sensitivity values 0.05 and 0.15, as a practical floor on a regression worth reporting: a ten-point drop in pass rate, 5 of 50 hand-authored tasks or 9 of 90 generated scenarios. It is a convention, not a quantity derived from the data, so \S\ref{sec:ceiling} reports every conclusion that depends on it as a function of $\theta$. Suites: \textbf{S1}, the hand-authored tasks; \textbf{S2}, the generated suite subsampled to $|\text{S1}|$ over 20 draws (detected if a majority detect); \textbf{S3}, the full generated suite. Each scenario runs once per configuration.

\textbf{Coverage.} For each mutation we compute mechanically, from the archive's own descriptors, the set $C_m$ of scenarios exercising the mutated element: for S3, those whose recorded requirement is the one mutated or whose grading names the affected tool; for S1, tasks whose criteria or instruction involve it. $C_m$ is knowable \emph{before any conversation runs}.

\textbf{The ceiling.} A defect can only change a scenario it reaches, so if $b$ of the $|C_m|$ covered scenarios pass at baseline, the most it can itself contribute to the suite-level drop over a suite of $N$ is every one of those $b$ flipping: $\mathrm{ceil}(m,\Sigma) = b/N$. Since $b$ is read off the baseline pass outcomes, the ceiling is computable from baseline results alone, before the defect is introduced. When $\mathrm{ceil}(m,\Sigma) \le \theta$, the within-footprint contribution to $\Delta_m$ cannot exceed $\theta$; an observed $\Delta_m$ can still exceed the ceiling, through run-to-run noise or through behavioural effects the trace does not capture. The bound assumes the trace captures the affected set; for T2 it does not, since the renamed read tool is named in no grading criteria. \textbf{Post-hoc scores}, defined after those results and labelled as such throughout: \emph{scoped} detection applies the same test to $C_m$ alone; \emph{grading views} recompute $\Delta_m$ under programmatic-only and judge-only grading.

\section{Results}

\subsection{The ceiling is binding}
\label{sec:ceiling}

Under the specified score the generated suite detects 1 of 7 airline defects and the hand-authored suite 0 of 7, as does the size-matched S2; baseline pass rates 0.50 (S1) and 0.88 (S3). Table~\ref{tab:ceiling} shows most of that outcome was bounded before the defects were introduced: of the 5 airline defects with a traceable footprint, \textbf{4 have a ceiling at or below $\theta$ on both S1 and S3}. R1's S3 coverage is zero scenarios, so its ceiling is exactly zero: the archive has no cell for the deleted clause, and a requirement-traced suite can report that about itself; a hand-authored suite cannot.

\textbf{Sensitivity to $\theta$.} Both halves of this outcome move with $\theta$, and we report them over its whole range (Appendix~\ref{app:theta}). Since the interval condition does not depend on $\theta$, each detection holds for every $\theta$ below its $\Delta$: airline O1 on S3 below 0.14, retail O1 on S1 below 0.17, and T2 on airline S3 joins below 0.10; no airline S1 or S2 detection appears at 0.05, 0.10 or 0.15. The ceiling count is \emph{not} robust to lowering $\theta$: 4 of 5 traceable defects are ceiling-bound for every $\theta$ from 0.10 up to C1's ceiling of 0.49, but 3 at 0.09, 2 at 0.08 and only R1 at 0.05. Lowering $\theta$ frees R2 and T1 from the ceiling yet detects neither, because their whole-suite intervals do not exclude zero; the one added detection is T2, whose affected set the trace does not capture.

\begin{table}[t]
\centering
\footnotesize
\setlength{\tabcolsep}{4.5pt}
\renewcommand{\arraystretch}{0.9}
\caption{Coverage, ceiling, and what each score sees. $|C_m|/N$ is the traced footprint over suite size, \emph{ceil} is $b/N$, positive $\Delta$ a drop in pass rate, $\dagger$ a detection, ``n/a'' a coverage hole; remaining rows in Appendices~\ref{app:all} and~\ref{app:retail}.}
\label{tab:ceiling}
\begin{tabular}{@{}llrrcc@{}}
\toprule
 & & $|C_m|/N$ & ceil & whole-suite $\Delta$ [95\% CI] & scoped $\Delta$ [95\% CI] \\
\midrule
\multicolumn{6}{@{}l}{\emph{airline}} \\
R1 & S1 & 13/50 & 0.04 & -0.06 [-0.22, +0.10] & -0.31 [-0.54, -0.08] \\
R1 & S3 & \phantom{0}0/90 & 0.00 & +0.04 [-0.03, +0.13] & n/a \\
T1 & S1 & \phantom{0}7/50 & 0.06 & -0.04 [-0.18, +0.10] & +0.43 [+0.14, +0.86]$^\dagger$ \\
T1 & S3 & 12/90 & 0.09 & +0.08 [\phantom{+}0.00, +0.17] & +0.67 [+0.42, +0.92]$^\dagger$ \\
O1 & S3 & \phantom{0}9/90 & 0.10 & +0.14 [+0.04, +0.26]$^\dagger$ & +0.11 [\phantom{+}0.00, +0.33] \\
C1 & S3 & 51/90 & 0.49 & +0.02 [-0.07, +0.10] & +0.08 [-0.04, +0.20] \\
\multicolumn{6}{@{}l}{\emph{retail}} \\
O1 & S1 & \phantom{0}0/48 & 0.00 & +0.17 [+0.04, +0.31]$^\dagger$ & n/a \\
T1 & S1 & 14/48 & 0.13 & -0.02 [-0.21, +0.17] & +0.43 [+0.21, +0.71]$^\dagger$ \\
\bottomrule
\end{tabular}
\end{table}

\textbf{A defect can saturate its ceiling and still read as an improvement.} Removing \texttt{cancel\_reservation} (T1) fails \emph{every} hand-authored task whose checks require it (3 of 3 passing at baseline on airline, 6 of 6 on retail): exactly the ceiling. Yet the airline pass rate moves from 25 to 27 of 50 tasks and retail's from 20 to 21 of 48: off-target drift exceeds what the defect destroyed. In both domains the score reports a total, verified tool breakage as a small \emph{improvement}: averaging over the archive does not merely dilute the signal, it inverts it. C1 is the complement: the study's highest coverage (51 of 90, ceiling 0.49) and still no detection, so a generous ceiling is necessary, not sufficient.

\subsection{Two regimes}
\label{sec:regimes}

\textbf{Trace-aligned defects} (R1, R2, T1) have footprints the generator's descriptors express: a requirement, a tool. Coverage is small and the ceiling binds; scoring at the covered cells recovers T1 in full: scoped $\Delta$ = +0.43 (airline S1), +0.67 (airline S3) and +0.43 (retail S1), every interval excluding zero. Three reruns of each covered airline S1 task, run after the main analysis as a diagnostic, qualify the size but not the breakage: the mutant passes none of 21 trials, while the baseline passes only 3 of 21, all on one task, so two of the three single-run baseline passes did not recur and the rerun scoped $\Delta$ is 0.14. Recovery is not automatic: R2's scoped $\Delta$ reaches $\theta$ on neither suite (Appendix~\ref{app:all}), and R1's is a statistically significant scoped \emph{improvement} ($\Delta$ = -0.31, interval excluding zero): the mutant is better on the 13 tasks it touches, because the baseline agent was over-refusing and the defect removed the clause it over-applied. A suite-level difference sums regressions and accidental fixes; per-cell scoring separates them.

\textbf{Trace-orthogonal defects} are the other regime; O1 is the clean case. Forcing escalation raises the share of airline hand-authored conversations containing a transfer from 36\% to 84\%, but that footprint is behavioural, not requirement-shaped, and the archive has no descriptor for it. Only 9 of 90 airline generated scenarios name escalation; on retail the generator extracted \emph{no} escalation requirement and none of the five transfer-naming tasks fell in the seeded subsample, so traced coverage is zero on both suites. Coverage predicts nothing here: at a ceiling of exactly zero, retail O1 is detected on S1 at $\Delta$ = +0.17 [+0.04, +0.31], and airline O1 on S3 is the study's only specified-score detection, and a threshold-fragile one, absent at $\theta$ = 0.15 (Appendix~\ref{app:all}). For these defects the whole suite is already the right unit, and scoping does \emph{not} help: airline O1's scoped interval touches zero on both suites. The regime also decides which grading view can see the defect: under programmatic end-state grading O1 scores no worse (airline S3 $\Delta$ = +0.07, retail S3 $\Delta$ = 0.00), because where the expected end state is reachable by inaction an end-state check cannot tell doing nothing from doing the right thing; the transcript judge sees it in both domains ($\Delta$ = +0.22 [+0.10, +0.36] airline, +0.20 [+0.06, +0.34] retail).

\subsection{Where diversity buys detection}
\label{sec:diversity}

Both ablations reuse the conversations already run (Appendix~\ref{app:ablation}). \textbf{Scenario kind, at fixed suite size, is the one size-matched diversity contrast here, and it is positive}, though exploratory, since all-kind draws also shift requirement composition, so kind is not varied alone. Holding the airline suite at 30 scenarios, a happy-path-only suite detects none of the 7 defects, while kind-diverse suites of the same size detect 6 times across 140 draw-by-mutation trials, 5 of them O1, detected in 5 of 20 draws. The gain falls where the two-regime analysis predicts: edge and adversarial-benign scenarios provoke the escalation path, supplying the behavioural coverage the requirement descriptor cannot express. \textbf{Persona count is inconclusive.} Detection rises with persona count then falls at the maximum in both domains, but sub-suite size grows with persona count, confounding size with diversity; we claim no curve, only that turning the persona dial up does not monotonically buy detection.

\section{Implications}

\emph{Score the archive per cell, not by its mean.} A generated suite already records which requirement each scenario exercises; a suite-level pass rate discards that and, as \S\ref{sec:ceiling} shows, can invert the sign of a total breakage. \emph{Publish coverage as a first-class suite property.} The ceiling needs only baseline results, so a suite can state before a defect is introduced where its own scoring is weakest. \emph{Match the descriptor space to the defect space.} Requirements and personas do not span the behaviours a defect can change; a behavioural descriptor (does this conversation escalate, does it call this tool) would have given O1 a cell in both domains, making quality-diversity search over the \emph{defect} space the natural next step.

\textbf{Designing and learning behavioural descriptors.} A behavioural descriptor is a function of the trajectory, not of the specification, so it is computed from transcripts rather than read off the generator. \emph{Designed} descriptors come from the agent's action vocabulary, which its tool schemas and policy already enumerate: each tool call, each terminal action (write, refuse, escalate) and the read/write split gives a binary cell that a deterministic trace parser evaluates on baseline transcripts. Coverage and the ceiling carry over unchanged, with $C_m$ the scenarios whose baseline trajectory occupies the affected cell, still computable before the defect is introduced. O1's footprint is such a cell: the share of airline hand-authored conversations that escalate moves from 36\% to 84\%. \emph{Learned} descriptors serve where no single action names the failure: cluster baseline trajectories by tool-call sequence and use cluster membership as the cell, so a defect appears as mass moving between cells, testable with the same paired bootstrap on occupancy instead of pass rate and without a grading predicate. An occupancy shift flags changed behaviour, not by itself a regression; it says where to grade. Two cautions follow from our data. A cell the baseline rarely visits has a small ceiling of its own, so behavioural cells must be filled by search that targets them; edge and adversarial-benign scenarios already do part of this (\S\ref{sec:diversity}). And the transcript judge partly sees behaviour already: its strictness concentrates on O1 in both domains (Appendix~\ref{app:grader}), so judge-check disagreement is a cheap pointer to a missing behavioural cell. We have not evaluated either construction here.

\section{Limitations and conclusion}

We study two domains of one benchmark family and one agent model; the user simulator and judge are that same model, so all three may share blind spots, and judge-programmatic agreement is modest (Appendix~\ref{app:grader}). Pass rates carry run-to-run noise the bootstrap only partly captures; on S1 intervals are about $\pm$0.15 wide, as large as $\theta$, limiting power independently of the ceiling. Every operator is instantiated once per domain; T1 and O1 have a second instance on retail, with a different tool and policy, and each falls in the same regime as on airline, but no result generalises to its operator class; two to three instances per operator, with repeated rollouts, are the validation this pilot most needs. The retail replication covers two of the seven operators and was run after the airline results were known. Scoped scores, grading views and both ablations are post hoc on small subsets, and the ablation counts are draws, not independent samples.

Reading suite generation as diversity-driven search makes a practical prediction, and we confirm it: for defects whose affected set the trace captures, baseline coverage bounds what the within-footprint effect can contribute to a suite-level score, that bound is computable from baseline results alone, and at the specified threshold it already binds for most such defects. The fix is not more scenarios; it is scoring the archive the way archives are meant to be scored.

\bibliographystyle{plainnat}
\bibliography{refs}

\clearpage
\appendix
\section{Mutation operators}
\label{app:mutations}

\begin{table}[h!]
\centering
\footnotesize
\setlength{\tabcolsep}{4pt}
\renewcommand{\arraystretch}{0.95}
\caption{Mutation operators, one instance each; \emph{field grounding} names the failure class each models.}
\label{tab:mutations}
\begin{tabular}{@{}ll>{\raggedright\arraybackslash}p{6.0cm}>{\raggedright\arraybackslash}p{4.0cm}@{}}
\toprule
Id & Class & Operator & Field grounding \\
\midrule
R1 & Requirement & Delete one policy clause (24-hour cancellation) & specification violation~\citep{cemri2025mast} \\
R2 & Requirement & Invert one policy condition (``cannot'' $\rightarrow$ ``can'') & incorrect functionality~\citep{zhu2026where} \\
T1 & Tool & Remove one tool (\texttt{cancel\_reservation}) & tool-contract breakage~\citep{asgari2025challenges} \\
T2 & Tool & Rename one parameter in a tool schema & API/serialization mismatch~\citep{zhu2026where} \\
O1 & Orchestration & Invert the escalation rule (always transfer to a human) & orchestration and execution control~\citep{asgari2025challenges} \\
C1 & Context & Truncate the policy document to its first 40\% & context loss~\citep{asgari2025challenges,cemri2025mast} \\
C2 & Context & Truncate conversation history to the last 4 turns & loss of history~\citep{cemri2025mast} \\
\bottomrule
\end{tabular}
\end{table}

\section{Grader validity}
\label{app:grader}

On the 488 dual-graded airline transcripts (the 61 generated scenarios carrying both a programmatic check and a judged predicate, over baseline and seven mutants) the two signals agree on 75\% of verdicts ($\kappa$ = 0.21); the judge is lenient on 10\% and strict on 16\%. Its strictness concentrates on O1: it fails 26\% of O1 conversations the programmatic check passes, against 13\% at baseline and 14\% for the other six mutations, which is the same trace-orthogonality of \S\ref{sec:regimes} reappearing in the grader. Retail shows the same concentration on its 33 dual-graded S3 scenarios: the judge fails 30\% of O1 conversations the programmatic check passes, against 12\% at baseline and 12\% for T1.

\section{Whole-suite results for every airline mutation}
\label{app:all}

\begin{table}[h!]
\centering
\footnotesize
\setlength{\tabcolsep}{5pt}
\caption{Whole-suite score as specified before running experiments, airline, all seven mutations: $\Delta_m$ with paired-bootstrap 95\% interval. S2 values are means over the 20 size-matched draws. Detection requires $\Delta_m > \theta = 0.10$ and an interval excluding zero; the only detection is marked $\dagger$. S2 is detected only if a majority of draws detect, so its mean interval can exclude zero without a detection. T2 on S3 equals $\theta$ exactly (9 of 90 scenarios) and so does not exceed it. Baseline pass rates 0.50 (S1), 0.88 (S2 mean), 0.88 (S3). The single detection is threshold-fragile: at $\theta$ = 0.05 T2 is also detected on S3, and at $\theta$ = 0.15 nothing is detected by any suite.}
\label{tab:all}
\begin{tabular}{@{}lccc@{}}
\toprule
Mutation & S1 & S2 & S3 \\
\midrule
R1 & -0.06 [-0.22, +0.10] & +0.05 [-0.05, +0.16] & +0.04 [-0.03, +0.13] \\
R2 & +0.06 [-0.10, +0.22] & \phantom{+}0.00 [-0.10, +0.10] & \phantom{+}0.00 [-0.08, +0.08] \\
T1 & -0.04 [-0.18, +0.10] & +0.08 [-0.02, +0.19] & +0.08 [\phantom{+}0.00, +0.17] \\
T2 & +0.12 [-0.02, +0.26] & +0.10 [-0.01, +0.22] & +0.10 [+0.01, +0.19] \\
O1 & -0.02 [-0.16, +0.12] & +0.15 [+0.02, +0.28] & +0.14 [+0.04, +0.26]$^\dagger$ \\
C1 & +0.14 [\phantom{+}0.00, +0.28] & +0.02 [-0.10, +0.13] & +0.02 [-0.07, +0.10] \\
C2 & -0.08 [-0.22, +0.06] & -0.04 [-0.14, +0.05] & -0.04 [-0.12, +0.03] \\
\bottomrule
\end{tabular}
\end{table}

R2's traced coverage on S3 is 3 of 90 scenarios (ceiling 0.02) and on S1 15 of 50 tasks (ceiling 0.08); both lie below $\theta$, which with R1, T1 and O1 gives the four ceiling-bound defects of \S\ref{sec:ceiling}; O1 qualifies on a ceiling of exactly $\theta$ on both suites, and its footprint is in any case only incompletely traced (\S\ref{sec:regimes}). R2's scoped $\Delta$ is +0.07 [-0.13, +0.27] on S1 and 0.00 [0.00, 0.00] on its 3 S3 scenarios; T2's is +0.12 [-0.08, +0.32] on its 25 S1 tasks, coinciding with its whole-suite value. O1's S1 row, omitted from Table~\ref{tab:ceiling} for space, has traced coverage 8 of 50 (ceiling 0.10), whole-suite $\Delta$ = -0.02 and scoped $\Delta$ = -0.12 [-0.38, 0.00], the scoped interval touching zero, as \S\ref{sec:regimes} notes. C2 has no principled subset (history truncation touches every conversation long enough to lose a turn), so no ceiling is computed for it.

\section{Retail replication}
\label{app:retail}

The airline protocol was rerun unchanged on the retail domain of \taubench{} for the baseline and two mutations, O1 (invert the escalation rule) and T1 (remove the tool \texttt{exchange\_\allowbreak delivered\_\allowbreak order\_\allowbreak items}), after the airline results were known; it is a post-hoc replication, not a study specified in advance over the operator set. Suites: S1, 48 hand-authored tasks drawn as a seeded subsample of the domain's 114 (two conversations were unrecoverable); S2, 20 size-matched draws; S3, a seeded 50-scenario subsample of the 90 generated, 33 of which carry a programmatic end-state check. Baseline pass rates 0.42 (S1), 0.86 (S2 mean), 0.86 (S3). All but three of retail's hand-authored tasks carry no judged predicate, so we report no S1 judge-only view. No S3 scenario names the removed tool in its grading criteria, so T1 has no traced S3 subset: the same trace limit as T2 on airline. Judge-programmatic agreement on the 99 dual-graded S3 transcripts is 74\% ($\kappa$ = 0.09), lower than on airline and on a smaller overlap.

\begin{table}[h!]
\centering
\footnotesize
\setlength{\tabcolsep}{5pt}
\caption{Retail replication, airline protocol unchanged: $\Delta_m$ with paired-bootstrap 95\% interval, positive is a drop in pass rate; combined grading unless a view is named; S2 is the mean over 20 draws. Detection as in Table~\ref{tab:all}; scoped rows are post hoc, $n$ is the number of scenarios exercising the mutated element, flips are pass$\to$fail\,/\,fail$\to$pass.}
\label{tab:retail}
\begin{tabular}{@{}llrcc@{}}
\toprule
Mutation & Suite & $n$ & $\Delta$ [95\% CI] & detected \\
\midrule
O1 & S1 & 48 & +0.17 [+0.04, +0.31] & yes \\
O1 & S2 & 48 & +0.10 [-0.04, +0.24] & no \\
O1 & S3 & 50 & +0.10 [-0.04, +0.24] & no \\
O1 & S3, programmatic only & 33 & \phantom{+}0.00 [-0.12, +0.12] & no \\
O1 & S3, judge only & 50 & +0.20 [+0.06, +0.34] & yes \\
T1 & S1 & 48 & -0.02 [-0.21, +0.17] & no \\
T1 & S1, scoped (flips 6/0) & 14 & +0.43 [+0.21, +0.71] & yes \\
T1 & S2 & 48 & +0.02 [-0.11, +0.15] & no \\
T1 & S3 & 50 & +0.02 [-0.10, +0.16] & no \\
\bottomrule
\end{tabular}
\end{table}

\section{Ablation protocol and results}
\label{app:ablation}

Both ablations reuse the conversations already run: no additional simulation was performed, and each sub-suite is scored with the same $\theta$, the same 2,000-sample paired bootstrap and the same seed as the main analysis.

\begin{table}[h!]
\centering
\footnotesize
\setlength{\tabcolsep}{5pt}
\renewcommand{\arraystretch}{0.95}
\caption{Diversity ablations on the generated suite, counted as detections over (draw $\times$ mutation) trials at $\theta = 0.10$. Airline instantiates 7 mutations, retail 2. \emph{Scenario kind} is size-matched: the happy-only arm is the single deterministic sub-suite of all happy scenarios, the all-kinds arm is 20 draws of the same size from all three kinds. \emph{Personas} is not size-matched (Appendix~\ref{app:ablation}). Retail's kind ablation is degenerate and reported for completeness.}
\label{tab:diversity}
\begin{tabular}{@{}llcc@{}}
\toprule
 & & airline & retail \\
\midrule
\multicolumn{2}{@{}l}{\emph{Scenario kind} (size-matched)} & size 30 & size 16 \\
& happy path only & 0 / 7 & 0 / 2 \\
& all three kinds & 6 / 140 \quad(O1: 5 of 20 draws) & 0 / 40 \\
\midrule
\multicolumn{2}{@{}l}{\emph{Personas}} & \multicolumn{2}{c}{(not size-matched)} \\
& $P = 1$ & \phantom{0}2 / 140 \quad(size 11) & 2 / 40 \quad(size \phantom{0}6) \\
& $P = 2$ & \phantom{0}6 / 140 \quad(size 23) & 4 / 40 \quad(size 12) \\
& $P = 4$ & 22 / 140 \quad(size 45) & 4 / 40 \quad(size 25) \\
& $P = 8$ & 11 / 140 \quad(size 50) & 0 / 40 \quad(size 48) \\
\bottomrule
\end{tabular}
\end{table}

\emph{Scenario kind.} The happy-only arm is the full set of happy-path generated scenarios: a single deterministic sub-suite, hence 1 trial per mutation. The all-kinds arm draws 20 sub-suites of that same size uniformly from all three kinds. Suite size is therefore matched between arms, but kind composition is not the only thing that varies with it: drawing uniformly from all three kinds also changes which requirements the sub-suite covers.

\emph{Personas.} For each level $P$ we draw 20 times: sample $P$ distinct personas, then sample $\min(|\text{S1}|, \text{scenarios available})$ scenarios from the union of their scenarios. Because a persona carries roughly $90/8$ scenarios, the available pool (and hence sub-suite size) grows with $P$ until it saturates at $|\text{S1}|$. Size and diversity are therefore confounded, and at $P = 4$ the sample is very nearly the whole union, so consecutive draws differ only in persona choice. We report the counts and draw no curve from them.

\section{Threshold sensitivity}
\label{app:theta}

All values are recomputed from the frozen results with no additional runs (script released with the pipeline). For S1 and S3, detection at any $\theta$ follows from the stored $\Delta$ and interval; the recomputation reproduces the stored detections at all three specified values exactly. S2 detection is a majority over 20 draws whose per-draw $\Delta$ is not stored, so S2 is reported at the three specified values only. A traceable defect is \emph{ceiling-bound} if its ceiling is at or below $\theta$ on every suite where it has a traced subset; airline ceilings are R1 0.04 (S1) and 0.00 (S3), R2 0.08 and 0.02, T1 0.06 and 0.09, O1 0.10 and 0.10, C1 0.49 (S3 only).

\begin{table}[h!]
\centering
\footnotesize
\setlength{\tabcolsep}{5pt}
\caption{Specified-score detections and ceiling-bound airline defects at the three thresholds fixed before the runs. ``none'' means no mutation is detected on that suite.}
\label{tab:theta}
\begin{tabular}{@{}lcccccc@{}}
\toprule
 & \multicolumn{3}{c}{airline detections} & \multicolumn{2}{c}{retail detections} & ceiling-bound \\
\cmidrule(lr){2-4}\cmidrule(lr){5-6}
$\theta$ & S1 & S2 & S3 & S1 & S2, S3 & (airline, of 5) \\
\midrule
0.05 & none & none & O1, T2 & O1 & none & 1 (R1) \\
0.10 & none & none & O1 & O1 & none & 4 (R1, R2, T1, O1) \\
0.15 & none & none & none & O1 & none & 4 (R1, R2, T1, O1) \\
\bottomrule
\end{tabular}
\end{table}

Over a continuous grid the ceiling-bound count is 0 below 0.04, 1 from 0.04, 2 from 0.08, 3 from 0.09 (T1's S3 ceiling is 0.089), 4 from 0.10 and 5 from C1's 0.49.

\section{Run configuration}
\label{app:repro}

Both domains ran on 29 August 2026 in the unmodified \taubench{} harness, v1.0.1 (commit fc0055d): airline from 09:04 to 12:59 UTC, retail from 13:35 to 16:11 UTC. Agent, user simulator and judge are \texttt{gpt-5.6-luna}; the generator is \texttt{gpt-5.6-terra}. Temperature is omitted, so the provider default applies, because the agent model rejects temperature 0; reasoning effort is set to none, and the usage records show zero reasoning tokens per conversation. Each scenario runs once per configuration, at most two conversations concurrently. Analysis uses a 2,000-sample paired bootstrap over scenarios with a fixed seed and 20 S2 draws. C1 keeps the first 3,070 of the airline policy's 7,676 characters; C2 keeps the last 4 turns. Airline simulation and judging cost \$3.17 over 1,120 conversations, retail \$1.54 over 300, and generating both suites \$2.31.

\end{document}